\documentclass{article}

\usepackage[utf8]{inputenc}
\usepackage{graphicx}

\usepackage[preprint, nonatbib]{neurips_2024}

\usepackage[utf8]{inputenc} 
\usepackage[T1]{fontenc}    
\usepackage{hyperref}       
\usepackage{url}            
\usepackage{booktabs}       
\usepackage{amsfonts}       
\usepackage{nicefrac}       
\usepackage{microtype}      
\usepackage{xcolor}         

\usepackage{colortbl}
\usepackage{amsmath}
\usepackage{subfigure}

\title{PoinTramba: \\ \vspace{-7mm} \includegraphics[width=0.1\textwidth]{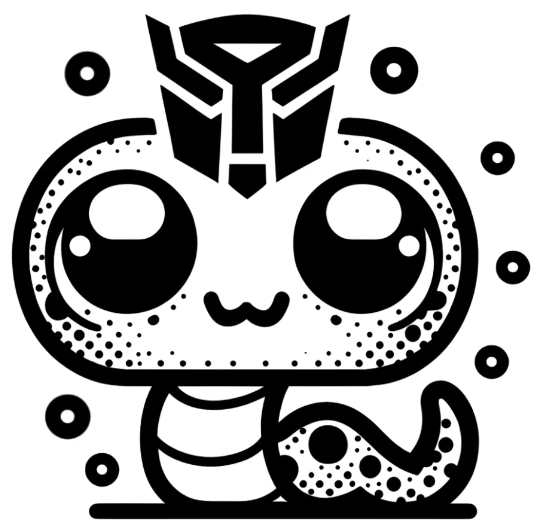} A Hybrid Transformer-Mamba Framework \\ for Point Cloud Analysis}

\author{
  Zicheng Wang$^{1, 2}$, Zhenghao Chen$^2$, Yiming Wu$^1$, Zhen Zhao$^2$, Luping Zhou$^2$, Dong Xu$^1$\thanks{Corresponding author} \\
  $^1$The University of Hong Kong \\
  $^2$The University of Sydney \\
  \texttt{\{edmond02, yimingwu, dongxu\}@hku.hk} \\
  \texttt{\{zhenghao.chen, zhen.zhao, luping.zhou\}@sydney.edu.au} \\
}

\newcommand{\ie}{\textit{i.e.}}
\newcommand{\eg}{\textit{e.g.}}

\begin{document}

\maketitle

\begin{abstract}
Point cloud analysis has seen substantial advancements due to deep learning, although previous Transformer-based methods excel at modeling long-range dependencies on this task, their computational demands are substantial.
Conversely, the Mamba offers greater efficiency but shows limited potential compared with Transformer-based methods.
In this study, we introduce PoinTramba, a pioneering hybrid framework that synergies the analytical power of Transformer with the remarkable computational efficiency of Mamba for enhanced point cloud analysis. Specifically, our approach first segments point clouds into groups, where the Transformer meticulously captures intricate intra-group dependencies and produces group embeddings, whose inter-group relationships will be simultaneously and adeptly captured by efficient Mamba architecture, ensuring comprehensive analysis.
Unlike previous Mamba approaches, we introduce a bi-directional importance-aware ordering (BIO) strategy to tackle the challenges of random ordering effects. This innovative strategy intelligently reorders group embeddings based on their calculated importance scores, significantly enhancing Mamba's performance and optimizing the overall analytical process.
Our framework achieves a superior balance between computational efficiency and analytical performance by seamlessly integrating these advanced techniques, marking a substantial leap forward in point cloud analysis.
Extensive experiments on datasets such as ScanObjectNN, ModelNet40, and ShapeNetPart demonstrate the effectiveness of our approach, establishing a new state-of-the-art analysis benchmark on point cloud recognition. For the first time, this paradigm leverages the combined strengths of both Transformer and Mamba architectures, facilitating a new standard in the field.
The code is available at \url{https://github.com/xiaoyao3302/PoinTramba}.
\end{abstract}

\section{Introduction}
\label{intro}
Point clouds, which serve as crucial 3D visual signals containing essential geometric information, have garnered increasing research interest~\cite{PVRCNN, 3DSPS}. Point cloud analysis tasks, such as classification and segmentation, have achieved remarkable success thanks to the advent of deep learning methods. Early approaches either leveraged auxiliary data structures, like voxels, processed with 3D convolutional operations~\cite{VoxNet, ModelNet}, or directly utilized multilayer perceptions (MLPs)~\cite{PointNet, PointNet2, DGCNN} to extract visual representations from raw point sets. However, these methods primarily rely on local dependencies and often neglect global features.
%
%
To enhance global perception in point clouds, recent studies, \eg, PointBERT~\cite{PointBERT} and PointMAE~\cite{PointMAE}), have introduced Transformer~\cite{Transformer} with attention mechanisms to process long sequences.
Although the powerful long-range modeling capacity of the Transformer intuitively enhances analytical performance, the comprehensive mechanism inevitably results in high computational complexity and memory usage, rendering these methods less practical.  

On the other hand, to avoid the large memory burden when modeling long-range relationships, a new architecture called Mamba~\cite{Mamba} with a state space model~\cite{SSM, S4Model} module has been introduced. This architecture has achieved success in various Natural Language Processing~\cite{Mamba4Language, Densemamba} and Computer Vision~\cite{VMAMBA, VisionMAMBA} tasks. However, despite the efficiency of Mamba-based methods, their performance still lags behind that of Transformer-based methods of comparable size in point cloud analysis~\cite{PointMamba, PCMamba}. 
%
Furthermore, the application of Mamba to point cloud analysis remains an area requiring further investigation. For example, one significant challenge is that point clouds are inherently unordered, whereas Mamba is primarily designed to process structured data. Consequently, the effectiveness of Mamba in handling point cloud data is still uncertain.

In this study, we introduce \textbf{PoinTramba}, a novel hybrid framework for point cloud analysis that harnesses the robust analytical capabilities of Transformer along with the efficiency of Mamba.
Specifically, we segment point clouds into distinct groups and utilize the Transformer and Mamba to model intra-group and inter-group relationships, respectively.
Initially, capitalizing on the Transformer's exceptional ability to model long-range dependencies, we employ it to capture intra-group dependencies and generate enhanced intra-group features,\ie, group embeddings.
%
While using the Transformer to model intra-group point clouds remains feasible due to the small number of points within the group, this approach would lead to a significant increase in complexity when modeling inter-group dependencies, due to the large number of groups.
To efficiently model inter-group dependencies, we integrate Mamba, which reduces complexity to a linear scale. Specifically, we feed the group embeddings produced by the Transformer into a Mamba encoder. This encoder extracts inter-group global features from each point cloud sample, facilitating analytical tasks such as classification and segmentation.


Particularly, instead of applying Mamba directly to unordered point cloud groups, we introduce a bi-directional importance-aware ordering (BIO) strategy. This approach reorders the groups to mitigate the negative effects of random point cloud ordering to our Mamba encoder.
Unlike previous methods~\cite{PointMamba, PCMamba} such as those using conventional z ordering~\cite{z_order} or Hilbert ordering~\cite{Hilbert_order}, our proposed algorithm learns an ``importance'' score for each group embedding and reorders the groups accordingly.
%
Specifically, we map the group embeddings and the global features from the Mamba encoder to the same feature space. Then, we calculate the cosine similarity between each group embedding and the global feature. Additionally, we utilize an importance score prediction module that enables each group embedding to predict an importance score closely aligned with the calculated cosine similarity. Based on these importance scores, we reorder the group embeddings in a bi-directional manner, combining both descending and ascending orders. This ensures that each group embedding aggregates additional information from all other group embeddings.
%
This reordering strategy fully exploits Mamba's potential for processing structured data, resulting in more refined global inter-group features and significantly enhancing analytical performance.

%


By incorporating both Transformer and Mamba modules into our end-to-end hybrid analytical framework, we achieve comparable performance results on benchmark datasets while maintaining efficient complexity. Our contributions can be summarized as follows:
\textbf{1)} We propose PoinTramba, a novel hybrid framework combining Transformer and Mamba for efficient and effective point cloud analysis. This framework leverages the powerful modeling capacity of Transformers to produce enhanced intra-group features and the linear complexity of Mamba to generate inter-group features from a large number of group embeddings.
\textbf{2)} We propose a new bi-directional importance-aware ordering strategy to reorder the group embeddings. Such an operation can richly exploit the recurrent nature of Mamba
for better processing the structured group order and aggregating additional information from all other group embeddings.
\textbf{3)} We conduct extensive analytical experiments on point cloud classification and segmentation using ScanObjectNN, ModelNet40, and ShapeNetPart benchmark datasets. These experiments demonstrate that our method achieves comparable quantitative results and validates its effectiveness.

\section{Related Work}
\label{related_work}
\subsection{Deep Learning on Point Cloud Recognition}
Deep learning on point cloud recognition has attracted great attention as point clouds contain abundant depth information that can be used in various applications like autonomous driving and robotics~\cite{Transformer3DDet}. Compared with 2D images, 3D point clouds are sparse and the points are unordered, making it difficult to directly process point clouds for recognition.

Earlier analytical studies~\cite{VoxNet, ModelNet} have directly utilized auxiliary data structures like voxels, enabling the application of conventional deep learning methods such as 3D convolutions~\cite{3Dconv1, 3Dconv2}. However, using these additional data structures results in significant computational and memory costs.
Facing the issue, PointNet~\cite{PointNet} is a pioneering work that proposes to use MLPs to directly process the point sets, which is simple but achieves promising performance and has inspired a series of works like PointNet++~\cite{PointNet2} and DGCNN~\cite{DGCNN}, etc. However, these methods only rely on local dependencies that often neglect global features, thus leading to limited performance.
Recently, inspired by the success of the Transformer architecture in natural language processing and 2D vision on long-range context modeling~\cite{Transformer, ViT}, various works have been proposed to explore the effectiveness of the Transformer architecture in point clouds~\cite{PointTransformer, PCT, PointBERT, PointMAE} and have achieved great performance. 

However, the computational complexity of the attention module in Transformer is $\mathcal{O}(n^2)$, where $n$ indicates the number of input tokens. Therefore, despite the remarkable performance of the Transformer on point cloud recognition, when the number of input tokens increases, such Transformer-based methods will result in significant computational overhead, limiting the capacity of such methods.



\subsection{State Space Models}
The state space models are widely used in processing sequences of information by connecting the inputs and the outputs using latent states to model a system~\cite{MAMBA_survey}. Recently, inspired by the structured state space (S4) model~\cite{S4Model}, various works~\cite{S5Model, H3Model} have been proposed to use state space models to model long-range dependencies. In particular, the newly proposed Mamba~\cite{Mamba} has attracted great attention due to its great potential in global perception. Compared with the computational complexity of Transformer as $\mathcal{O}(n^2)$, the computational complexity of Mamba is $\mathcal{O}(n)$, leading to its linear scalability in sequence length. Inspired by the success of Mamba, various works have been proposed to examine the effectiveness of Mamba on 2D vision tasks, including image classification~\cite{VMAMBA, VisionMAMBA} and medical image segmentation~\cite{UMAMBA}, etc. However, these Mamba-based methods still cannot outperform Transformer-based methods with comparable size~\cite{JAMBA}. Therefore, various works have been proposed to combine the advantages of Transformer and Mamba for better performance and lower computational complexity, which is still an open issue and deserves exploring~\cite{JAMBA, MambaTransformer}.

\subsection{Mamba on Point Clouds}
Inspired by the success of Mamba on 2D vision tasks~\cite{VMAMBA, VisionMAMBA, UMAMBA}, some recent works have been proposed to examine the effectiveness of Mamba on 3D point clouds. In particular, PointMamba~\cite{PointMamba} combines the Mamba encoder with the standard point cloud group dividing operation while PCMamba~\cite{PCMamba} combines Mamba with PointMLP~\cite{PointMLP} to extract the potential of Mamba on point clouds. 

Compared with sequential language or image data, point clouds are highly unordered. However, Mamba is a recurrent model, and the order of the hidden states will severely influence the performance of the long-range dependency modeling of Mamba. Therefore, the key issue in adapting Mamba to point clouds lies in the ordering strategy of point clouds. To tackle the issue, PointMamba reorders the point clouds along the axes while PCMamba reorders the point clouds according to multiple orders including z order~\cite{z_order}, Hilbert order~\cite{Hilbert_order}, etc. However, whether these ordering strategies are suitable for point cloud analysis with Mamba is heuristic. 

\section{Method}
\label{method}


\subsection{PoinTramaba}
\label{PoinTramaba}
\begin{figure}[t]
\centering
\includegraphics[width=0.99\linewidth]{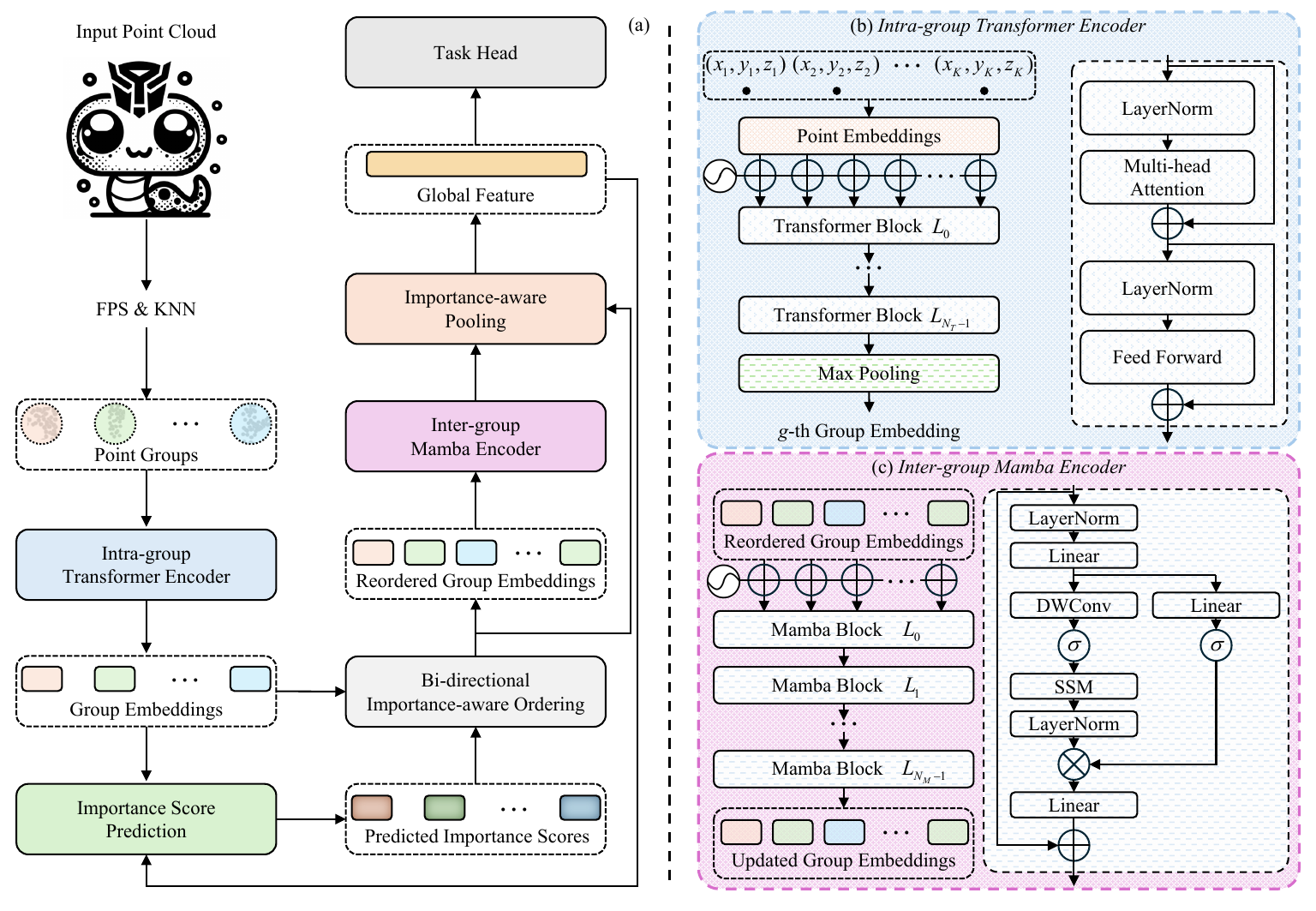}
\vspace{-1em}
\caption{The overview of our newly proposed PoinTramba framework (a) and its two main modules, the Intra-group Transformer Encoder (b) and the Inter-group Mamba Encoder (c). Initially, we segment the input point cloud into distinct point groups. Following this, we employ a Transformer encoder to model intra-group relationships and generate group embeddings. An importance-score prediction module is then utilized to predict the importance score for each group embedding. These predicted importance scores are used to reorder the group embeddings. Finally, a Mamba encoder extracts inter-group relationships from the reordered group embeddings, which are subsequently fed into an importance-aware pooling layer. This layer captures the global feature that can be further utilized for various downstream tasks such as classification and segmentation.}
\label{fig_pipeline}
\end{figure}

The pipeline of our method is shown in Fig.~\ref{fig_pipeline} (a), given an input point cloud  $\mathcal{P}$, we aim to produce a global feature $F$ that can be utilized for various downstream tasks.
Our method begins by segmenting $\mathcal{P}$ into $G$ point groups. For each group, we employ an \textit{Intra-group Transformer encoder}, consisting of $T$ Transformer layers, to generate $G$ group embeddings $\{\boldsymbol{e}_{g}\}_{g=1}^{G}$, where $g$ indicates the $g$-th point group, as shown in Fig.~\ref{fig_pipeline} (b).
Subsequently, we introduce a bi-directional importance-aware ordering (BIO) algorithm to reorder the group embeddings $\{\boldsymbol{e}_{g}\}_{g=1}^{G}$ in a bi-directional manner, combining both descending and ascending orders into $\{\boldsymbol{e}^{0}_{g}\}_{g=1}^{2G}$. 
These reordered embeddings are then passed through an \textit{Inter-group Mamba encoder}, as shown in Fig.~\ref{fig_pipeline} (c). After updating with $M$ Mamba layers, we obtain an updated set of features$\{\boldsymbol{e}^{M}_{g}\}_{g=1}^{2G}$, which is used to produce the global feature $\boldsymbol{f}$ via an importance-aware pooling (IAP) operation. Note that the superscript $m \in \{0, \cdots, M\}$ indicates the $m$-th Mamba layer, where $m=0$ indicates the input of the Mamba layer.

\subsection{Intra-group Transformer and Inter-group Mamba Encoder}
\label{Encoder}


\noindent \textbf{Intra-group Transformer Encoder.} 
We begin by segmenting the point cloud $\mathcal{P}$ into $G$ groups, each containing $K$ points, formally denoted as $\mathcal{P} = \{\mathcal{P}_{g}\}_{g=1}^{G}$, \textit{s.t.}, $\mathcal{P}_{g} = \{\mathcal{P}_{g_k}\}_{k=1}^{K}$. Initially, we employ the Farthest Point Sampling (FPS) algorithm to select $G$ keypoints. Subsequently, we use the K-Nearest Neighbors (KNN) algorithm to find the $K$ nearest neighbors for each keypoint. In this notation, the subscript $g$ refers to the $g$-th group of points, and $g_k$ denotes the $k$-th point within the $g$-th group.
%
%
%
After segmenting the point cloud into $G$ groups, we leverage the Transformer's exceptional capability to model long-range dependencies to capture intra-group dependencies and generate enhanced intra-group features, referred to as group embeddings. Specifically, for each point $\mathcal{P}_{g_k}$, we first project the coordinates of the point, \ie, $(x_{g_k}, y_{g_k}, z_{g_k})$, into a point embedding and a point position embedding using a standard embedding layer of the Transformer~\cite{Transformer, ViT, MAE}. The sums of these point embeddings and point position embeddings are then fed into a standard Transformer encoder, which consists of $T$ layers, to model intra-group relationships and generate the group embedding $\boldsymbol{e}_{g}$.

\noindent \textbf{Inter-group Mamba Encoder.} 
While using the Transformer to model intra-group point clouds is feasible due to the relatively small number of points within each group, modeling dependencies among a large number of groups (inter-group) would significantly increase computational complexity. 
%
To address this, we integrate the Mamba Encoder, which efficiently reduces the complexity to a linear scale when modeling inter-group dependencies.
Specifically, after obtaining the group embeddings $\{\boldsymbol{e}_{g}\}_{g=1}^{G}$, we propose a BIO strategy to obtain the reordered group embeddings $\{\tilde{\boldsymbol{e}}_{g}\}_{g=1}^{2G}$. More details about the BIO strategy will be provided in Sec.~\ref{BIO}.
Similarly, we reorder the position embeddings of the coordinates of the keypoints using our BIO strategy. With both reordered group and positional embeddings, we add them to obtain the aggregation $\boldsymbol{E}^{m}$ of the collection  $\{\boldsymbol{e}^{m}_{g}\}_{g=1}^{2G}$, which will be updated by an $M$-layer Mamba encoder.
As shown in Fig.~\ref{fig_pipeline} (c), we adopt the Mamba layer from~\cite{PointMamba}, which is a standard Mamba layer that can be presented as:

\begin{equation}
\begin{split}\label{eq4}
\boldsymbol{z}^m & =DW\_Conv\left(MLP\left(LN\left(\boldsymbol{E}^{m-1}\right)\right)\right), \\
\boldsymbol{E}^{m} & = MLP\left(LN\left(SSM\left(\sigma\left(\boldsymbol{z}^m\right)\right)\right) \cdot \sigma\left(L N\left(\boldsymbol{E}^{m-1}\right)\right)\right)+\boldsymbol{E}^{m-1}.
\end{split} 
\end{equation}

where $m$ indicates the $m$-th Mamba layer. Here, $DW\_Conv$ refers to depth-wise convolution, $LN$ denotes the LayerNorm operation, $SSM$ represents the state space model, which serves as a basic layer of our Mamba encoder, and $\cdot$ denotes the dot production. Additionally, $\sigma$ denotes the SiLU~\cite{SiLU} operation. After updating the group embeddings through $M$ Mamba layers, we obtain the final updated group embeddings $\{\boldsymbol{e}_{g}^{M}\}_{g=1}^{2G}$.
These updated group embeddings are subsequently passed into an importance-aware pooling (IAP) operation to extract the global feature $F$. By leveraging long-range relationship modeling and the linear scalability in sequence length, we enable the Mamba encoder to perform global perception.

\subsection{Importance-aware Ordering and Pooling}
\label{BIO}
\noindent \textbf{Bi-directional Importance-aware Ordering.} 
Due to the recurrent nature of Mamba~\cite{Mamba} and the unordered nature of point clouds, feeding randomly ordered group embeddings into the Mamba encoder significantly affects Mamba's performance of long-range dependency modeling. To address this issue, we propose a novel bi-directional importance-aware ordering (BIO) strategy to reorder the group embeddings, mitigating the adverse effects of random point cloud ordering.


\begin{figure}[t]
\centering
\includegraphics[width=0.99\linewidth]{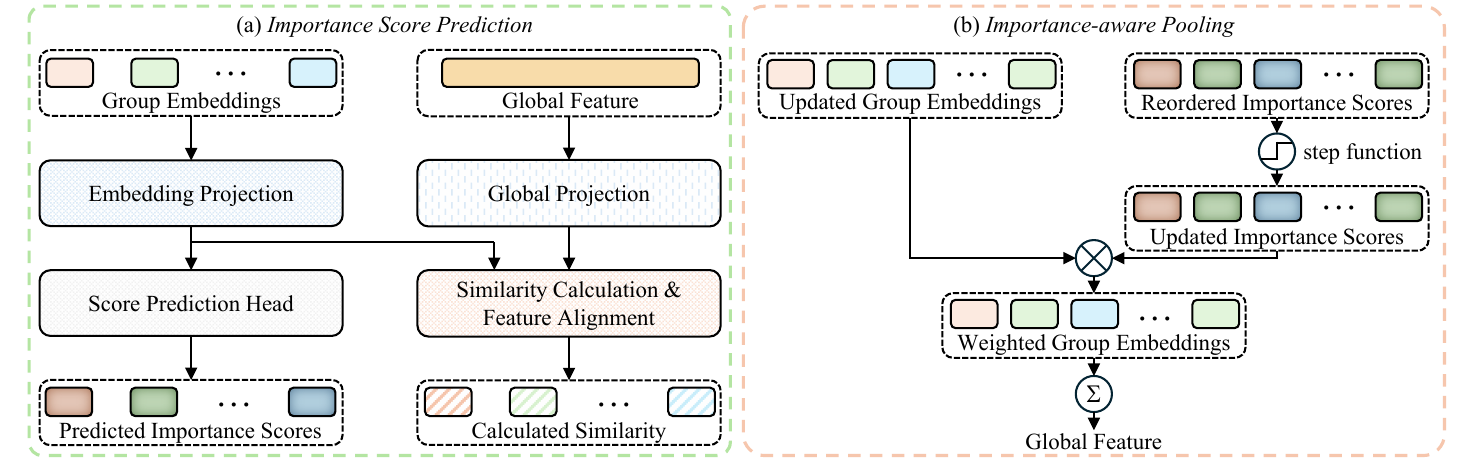}
\caption{The detailed design of our importance score prediction module (a) and our importance-aware pooling layer (b). The importance score prediction module targets at calculating the similarity between the group embeddings and the global feature, thus predicting the importance scores for group embeddings. The importance-aware pooling layer targets at aggregating the updated group embeddings to obtain the global feature.}
\vspace{-1em}
\label{fig_BIO}
\end{figure}

In particular, as illustrated in Fig.~\ref{fig_BIO}, given the group embeddings $\{\boldsymbol{e}_{g}\}_{g=1}^{G}$, we employ two non-linear projection layers, \ie, an embedding projection layer and a global projection layer. These two layers map the group embeddings and the global feature of the point cloud $\boldsymbol{f}$ into the same feature space, respectively, \ie, $\hat{\boldsymbol{e}}_{g}$ and $\hat{\boldsymbol{f}}$. Note we will introduce the global feature of the point cloud later. Therefore, we can calculate the cosine similarity $S_{g}$ between $\hat{\boldsymbol{e}}_g$ and $\hat{\boldsymbol{f}}$ as the importance score that can be presented as:
\begin{equation}
S_{g} = \frac{\hat{\boldsymbol{e}}_{g}^\top  \hat{\boldsymbol{f}}}{||\hat{\boldsymbol{e}}_{g}|| \times ||\hat{\boldsymbol{f}}||},
\end{equation}
However, calculating $S_{g}$ for each group embedding is not feasible as it requires a known global feature in the ordering stage, which is impractical. To overcome this limitation, we use a non-linear projection layer to predict the corresponding importance score of each group embedding, \ie, $I_{g}$. We encourage $I_{g}$ to approximate $S_{g}$ using an importance loss, $\mathcal{L}^{importance}$, which can be formulated as:
\begin{equation}
\mathcal{L}^{importance} = \frac{1}{N\times G} \sum_{n=1}^N \sum_{g=1}^G \mathcal{L}_{smooth}(S_{n, g}, I_{n, g}),
\end{equation}
where $\mathcal{L}_{smooth}$ is smooth $L1$ loss. 
Note that we introduce a new subscript $n$ here, indicating the $n$-th input point cloud, where there are $N$ point clouds in total. Note that below we may ignore the subscript $n$ when it is unnecessary to distinguish the index $n$ of the point cloud.

In addition, to learn meaningful projection layers, following previous works~\cite{DAS, PointGLR}, we perform a group embedding-to-global feature alignment using an alignment loss, \ie, $\mathcal{L}^{alignment}$, which can be formulated as:
\begin{equation}
\mathcal{L}^{alignment} = \frac{1}{N\times G} \sum_{n=1}^N \sum_{g=1}^G (-\mathrm{log} \frac{\hat{\boldsymbol{e}}_{n, g}^\top \hat{\boldsymbol{f}}_n}{\sum_{m}\hat{\boldsymbol{e}}_{n, g}^\top \hat{\boldsymbol{f}}_m}).
\end{equation}

After predicting the importance scores $I_{g}$ of the group embeddings, we reorder them in a bi-directional manner, \ie, $\boldsymbol{o}_{g} = \left[\boldsymbol{o}_{g}^1,\boldsymbol{o}_{g}^2\right]$, where $\boldsymbol{o}_{g}^1$ is the descending order of $I_{g}$ and $\boldsymbol{o}_{g}^2$ is the ascending order of $I_{g}$. This process yields the reordered group embeddings $\{\tilde{\boldsymbol{e}}_{g}\}_{g=1}^{2G}$. This bi-directional ordering strategy ensures that each group embedding aggregates information from all other group embeddings.

\noindent \textbf{Importance-aware Pooling.} 
Given the updated group embeddings $\{\boldsymbol{e}^{M}_{g}\}_{g=1}^{2G}$, we perform a pooling operation to obtain the global feature $\boldsymbol{f}$ from these embeddings. Recall the predicted importance scores of the group embeddings represent the cosine similarities between the group embeddings and the global feature, highlighting the importance of each group embedding. A negative importance score suggests that the corresponding group embedding negatively impacts the global feature. Therefore, we propose to discard group embeddings with negative importance scores and reweight the remaining embeddings using their importance scores. This process is represented as:
\begin{equation}
\boldsymbol{f} = \sum_{g=1}^{2G} \boldsymbol{e}^{M}_{g} \cdot \mathbf{1}(I_{g}),
\end{equation}
where $\mathbf{1}(\cdot)$ is the unit step function. The resulting global feature of the point cloud can be utilized for various downstream tasks such as classification or segmentation, using the corresponding loss $\mathcal{L}^{task}$.

\subsection{Objective Function}
\label{Loss}
We adopt a multi-faceted approach to loss computation for optimization. Specifically, we incorporate the following loss components: 
1) \textbf{Task loss} $\mathcal{L}^{task}$ optimizes the specific downstream tasks, such as classification or segmentation, ensuring robust performance. 2) \textbf{Importance loss} $\mathcal{L}^{Importance}$ optimizes the ordering of importance scores for different group embeddings. 3) \textbf{Alignment loss} $\mathcal{L}^{alignment}$ optimizes the prediction of importance scores, ensuring that the model learns semantically meaningful embedding projection layers. We use three trade-off parameters, $\alpha$, $\beta$, and $\gamma$, to balance each loss component. Consequently, we train the entire network by solving the following optimization problem in an end-to-end fashion:

\begin{equation}
\mathcal{L} = \alpha \mathcal{L}^{task} + \beta \mathcal{L}^{importance} + \gamma \mathcal{L}^{alignment}.
\end{equation}



\section{Experiments}
\label{exp}
\subsection{Experimental Protocols}
Following previous works~\cite{PointMamba}, we evaluate the effectiveness of our method on three downstream tasks, \ie, real-world object classification on ScanObjectNN~\cite{ScanObjNN}, synthetic object classification on ModelNet40~\cite{ModelNet}, and part segmentation on ShapeNetPart~\cite{ShapeNetPart}. The details of the datasets are provided in Sec.~\ref{dataset} and the implementation details are outlined in Sec.~\ref{implement}.


\begin{table}[t]
\caption{Comparison of classification accuracies (in \%) with the state-of-the-art methods on the ScanObjectNN on three variants, with PB-T50-RS being the most challenging one. The best performance is highlighted in bold. $^{\dagger}$ denotes using rotational augmentation for training. Hybrid denotes the hybrid Transformer and Mamba backbone. All of the methods are trained from scratch without pre-training.}
\centering
\scalebox{0.925}{
\begin{tabular}{l c c c c c}
\hline
Methods & Backbone & Param. (M) $\downarrow$ & OBJ-BG $\uparrow$ & OBJ-ONLY $\uparrow$ & PB-T50-RS $\uparrow$ \\
\hline
PointNet~\cite{PointNet} & MLP & 3.5 & 73.3 & 79.2 & 68.0 \\
PointNet++~\cite{PointNet2} & MLP & 1.5 & 82.3 & 84.3 & 77.9 \\  
PointCNN~\cite{PointCNN} & MLP & \textbf{0.6} & 86.1 & 85.5 & 78.5  \\
DGCNN~\cite{DGCNN} & MLP & 1.8 & 82.8 & 86.2 & 78.1 \\
MVTN~\cite{MVTN} & MLP & 11.2 & - & - & 82.8 \\
PointNeXt~\cite{PointNeXt} & MLP & 1.4 & - & - & 87.7 \\
PointMLP~\cite{PointMLP} & MLP & 13.2 & - & - & 85.4 \\
Point-BERT~\cite{PointBERT} & Transformer & 22.1 & 79.9 & 80.6 & 77.2 \\
PointMAE~\cite{PointMAE} & Transformer & 22.1 & 86.8 & 86.9 & 80.8 \\
PointMamba~\cite{PointMamba} & Mamba & 12.3 & 88.3 & 87.8 & 82.5 \\
PCM$^{\dagger}$~\cite{PCMamba} & Mamba & 34.2 & - & - & 88.1 \\
\rowcolor[HTML]{EFEFEF} 
PoinTramba (Ours) & Hybrid & 19.5 & \textbf{92.3} $\pm$ 0.4 & 90.9 $\pm$ 0.2 & 84.5 $\pm$ 0.1 \\
\rowcolor[HTML]{EFEFEF} 
PoinTramba$^{\dagger}$ (Ours) & Hybrid & 19.5 & \textbf{92.3} $\pm$ 0.2 & \textbf{91.3} $\pm$ 0.4 & \textbf{89.1} $\pm$ 0.2 \\
\hline
\end{tabular}
}
\label{table_ScanObjNN}
\end{table}

\begin{table}[t]
\caption{Comparison of classification accuracies (in \%) with the state-of-the-art methods on the ModelNet40. The best performance is highlighted in bold. $^{\ast}$ denotes reproduced results. $^{\ddagger}$ denotes using voting by averaging the results of 10 randomly scaled input point clouds. Hybrid denotes the hybrid Transformer and Mamba backbone. All of the methods are trained from scratch without pre-training.}
\centering
\scalebox{0.925}{
\begin{tabular}{l c c c}
\hline
Methods & Backbone & Param. (M) $\downarrow$ & Accuracy $\uparrow$ \\
\hline
PointNet~\cite{PointNet} & MLP & 3.5 & 89.2 \\
PointNet++~\cite{PointNet2} & MLP & 1.5 & 90.7 \\  
PointCNN~\cite{PointCNN} & MLP & \textbf{0.6} & 92.2  \\
DGCNN~\cite{DGCNN} & MLP & 1.8 & \textbf{92.9} \\
PointNeXt~\cite{PointNeXt} & MLP & 1.4 & \textbf{92.9} \\
OctFormer~\cite{OCTFormer} & Transformer & - & 92.7 \\
PointMAE~\cite{PointMAE} & Transformer & 22.1 & 92.3 \\
PointMamba~\cite{PointMamba} & Mamba & 12.3 & 92.4 \\
PCM$^{\ast}$~\cite{PCMamba} & Mamba & 34.2 & 92.6 \\
\rowcolor[HTML]{EFEFEF} 
PoinTramba (Ours) & Hybrid & 19.5 & 92.7 $\pm$ 0.1 \\
\rowcolor[HTML]{EFEFEF} 
PoinTramba$^{\ddagger}$ (Ours) & Hybrid & 19.5 & \textbf{92.9} $\pm$ 0.1 \\
\hline
\end{tabular}
}
\label{table_ModelNet}
\vspace{-0.4cm}
\end{table}

\subsection{Experimental results}
\label{result}
\noindent \textbf{Real-world Object Classification on ScanObjectNN.} We first compare our PoinTramba with other methods on the real-world object classification benchmark dataset ScanObjNN in Table~\ref{table_ScanObjNN}. It can be inferred from the table that our method surpasses the current methods, including the MLP-based methods like PointNet++ and Transformer-based methods like PointMAE, by a significant margin. Specifically, our method exceeds the previous state-of-the-art (SOTA) by 4.0\%, 3.5\% and 1.0\% on three variants of ScanObjNN, respectively. Notably, the number of parameters in PoinTramba is even less than those in standard Transformer-based methods, such as PointBERT and PointMAE. Although some MLP-based methods, like PointCNN, introduce fewer parameters, their performance lags significantly behind ours. The primary reason is that the intra-group Transformer encoder in our method enhances local feature extraction, outperforming other methods that rely on MLPs for this task. Additionally, the intra-group Transformer encoder is lightweight, ensuring that PoinTramba does not introduce significantly more parameters compared to PointMamba. It is also worth noting that PCM utilizes nearly double the parameters of PoinTramba, yet our method still outperforms it, highlighting the effectiveness of our approach.

\noindent \textbf{Synthetic Object Classification on ModelNet40.} We then compare our PoinTramba with other methods on the synthetic object classification benchmark dataset ModelNet40 in Table~\ref{table_ModelNet}. Note that we reproduce the result of PCM as it follows a different setting from ours. It can be inferred from the table that our method achieves the SOTA performance, surpassing the current Transformer-based and Mamba-based methods by a large margin.

\noindent \textbf{Part Segmentation on ShapeNetPart.} Finally, we compare our PoinTramba with other methods on the part segmentation benchmark dataset ShapeNetPart in Table~\ref{table_ShapeNetPart}. Note that we reproduce the results of PCM since it requires not only coordinates but also additional four-dimensional information as input. It can be inferred from the table that despite the ShapeNetPart being a highly competitive benchmark, our method achieves comparable performance to previous SOTA methods, validating the effectiveness of our approach.

\begin{table}[t]
\caption{Comparison of part segmentation accuracies (in \%) with the state-of-the-art methods on the ShapeNetPart. The mIoU for all instances (Inst.) is reported. The best performance is highlighted in bold. $^{\ast}$ denotes reproduced results. Hybrid denotes the hybrid Transformer and Mamba backbone. All of the methods are trained from scratch without pre-training. The inputs are 3D coordinates.}
\centering
\scalebox{0.925}{
\begin{tabular}{l c c c}
\hline
Methods & Backbone & Param. (M) $\downarrow$ & Inst. mIoU  $\uparrow$ \\
\hline
PointNet~\cite{PointNet} & MLP & - & 83.7 \\
PointNet++~\cite{PointNet2} & MLP & - & 85.1 \\  
DGCNN~\cite{DGCNN} & MLP & - & 85.2 \\
APES~\cite{APES} & MLP & - & \textbf{85.8} \\
PointMAE~\cite{PointMAE} & Transformer & 27.1 & 85.7 \\
PointMamba~\cite{PointMamba} & Mamba & \textbf{17.4} & \textbf{85.8} \\
PCM$^{\ast}$~\cite{PCMamba} & Mamba & 40.6 & 84.3 \\
\rowcolor[HTML]{EFEFEF} 
PoinTramba (Ours) & Hybrid & 25.4 & 85.7 $\pm$ 0.1 \\
\hline
\end{tabular}
}
\label{table_ShapeNetPart}
\end{table}


\begin{table}[t]
\caption{Ablation study on the effectiveness of different components in our method, including the intra-group Transformer encoder (Transformer), the inter-group Mamba encoder (Mamba), the alignment loss (Align), the bi-directional importance-aware ordering strategy (BIO) and the importance-aware pooling (IAP). Experiments are conducted on the PB-T50-RS variant of the ScanObjNN dataset. The baseline method for comparison (Variant No. 0) is PointNet++.}
\centering
\scalebox{0.925}{
\begin{tabular}{c | c c c c c | c}
\hline
Variant No. & Mamba & Transformer & Align & BIO & IAP & Acc. $\uparrow$ \\
\hline
0 & & & & & & 77.9 \\
1 & $\checkmark$ & & & & & 86.1 \\
2 & $\checkmark$ & & $\checkmark$ & & & 87.5 \\
3 & $\checkmark$ & $\checkmark$ & & & & 86.5 \\
4 & $\checkmark$ & $\checkmark$ & $\checkmark$ & & & 88.2 \\
5 & $\checkmark$ & & $\checkmark$ & $\checkmark$ & $\checkmark$ & 87.9 \\
6 & $\checkmark$ & $\checkmark$ & $\checkmark$ & $\checkmark$ & & 88.6 \\
7 & $\checkmark$ & $\checkmark$ & $\checkmark$ & $\checkmark$ & $\checkmark$ & 89.1 \\
\hline
\end{tabular}
}
\label{table_ablation_module}
\vspace{-0.4cm}
\end{table}

\subsection{Ablation Study}
\label{ablation}

In this section, we highlight the contributions of our module designs on the PB-T50-RS variant of the ScanObjNN dataset, as shown in Table~\ref{table_ablation_module}. From the table, we observe that deploying an inter-group Mamba encoder alone improves recognition performance by 8.2\% over PointNet++ (Variant No. 1 vs. Variant No. 0), primarily due to Mamba's global perception capability. Adding an intra-group Transformer encoder to the inter-group Mamba encoder further enhances performance by 0.4\% (Variant No. 3 vs. Variant No. 1), mainly owing to the Transformer's long-range modeling ability to capture intra-group dependencies. Incorporating our BIO strategy with the PoinTramba model results in a substantial performance improvement of 2.1\% (Variant No. 6 vs. Variant No. 3), underscoring the effectiveness of our ordering algorithm. Additionally, the importance-aware pooling operation helps the model focus on useful group embeddings while ignoring irrelevant ones, leading to a 0.5\% performance increase compared to the random ordering strategy (Variant No. 7 vs. Variant No. 6). Combining BIO and IAP with the Mamba encoder only can also achieve a recognition accuracy of 87.9\%, surpassing the Mamba encoder by 1.8\% (Variant No. 5 vs. Variant No. 1). Moreover, considering that the BIO strategy includes an alignment loss, which enhances local feature extraction, we further examine the effectiveness of $\mathcal{L}^{alignment}$. As shown in the table, adopting $\mathcal{L}^{alignment}$ alone improves the Mamba model's performance by 1.4\% (Variant No. 2 vs. Variant No. 1) and the PoinTramba model's performance by 1.7\% (Variant No. 4 vs. Variant No. 3). Additionally, the BIO strategy alone brings a 0.4\% performance improvement (Variant No. 4 vs. Variant No. 6). These ablation studies demonstrate the effectiveness of each component of our method.

\begin{figure}[t]
    \centering
    \subfigure[Ordering Strategies]{
        \includegraphics[width=0.475\textwidth]{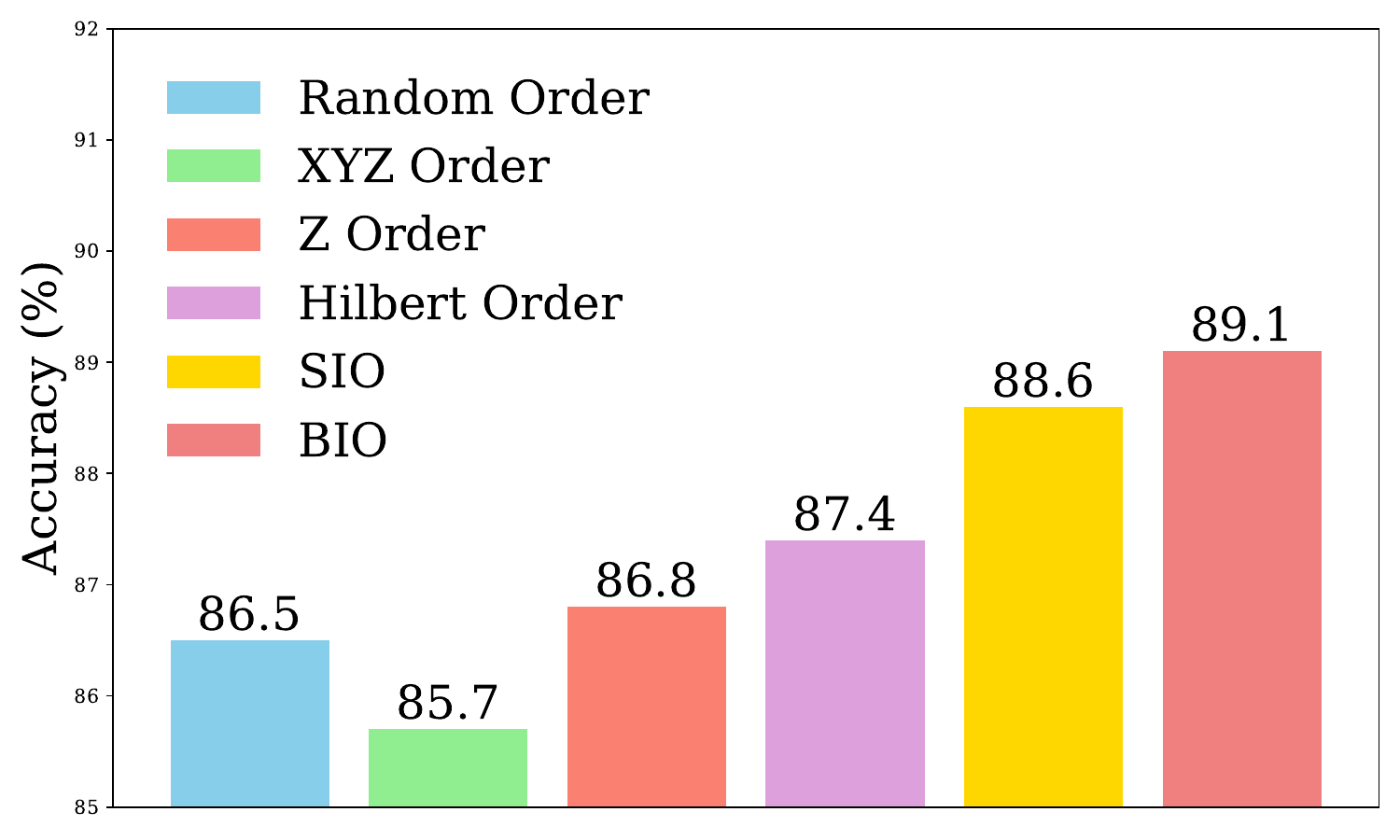}
    }
    \hfill
    \subfigure[Pooling Methods]{
        \includegraphics[width=0.475\textwidth]{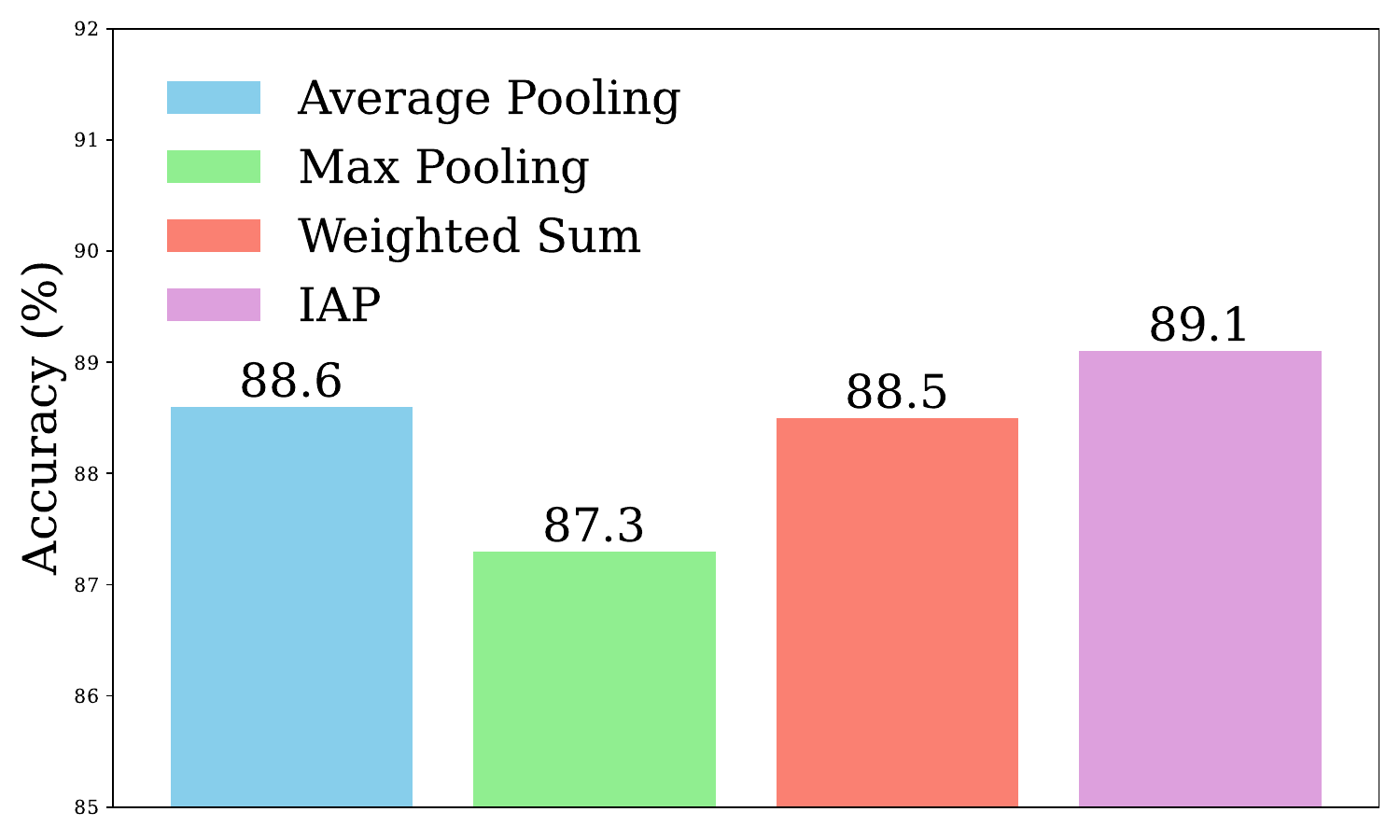}
    }
    \caption{Ablation studies on different ordering strategies and pooling methods. Experiments are conducted on the PB-T50-RS variant of the ScanObjNN dataset. PoinTramba is adopted as the backbone. (a) shows the comparison of different ordering strategies, \ie, random ordering strategy, XYZ ordering strategy, z ordering strategy, Hilbert Ordering strategy, our single-directional importance-aware ordering strategy (SIO) and bi-directional importance-aware ordering strategy (BIO). (b) illustrates the comparison of different pooling methods, \ie, average pooling, max-pooling, weighted sum and our importance-aware pooling (IAP).}
\label{fig_ablation2}
\end{figure}

We then verify the importance of ordering strategy for the Mamba model by adopting PoinTramba as the backbone and comparing six different ordering strategies, \ie, random ordering strategy, coordinate-based ordering strategy, \ie, the XYZ ordering strategy~\cite{PointMamba}, z ordering strategy~\cite{z_order}, Hilbert ordering strategy~\cite{Hilbert_order}, single-directional importance-aware ordering strategy in descending order (SIO) and bi-directional importance-aware ordering strategy (BIO). The results are shown in Fig.~\ref{fig_ablation2} (a). As can be seen from the table, the random ordering strategy outperforms the coordinate-based ordering strategy by 0.8\%. The main reason, based on our analysis, is likely that the random ordering strategy places embeddings that are both far from and close to a certain embedding into the neighboring region of that embedding. This way, the embedding aggregates information from both long-range and short-range distances, leading to competitive global perception. Notably, our BIO strategy surpasses the SIO strategy by 0.5\%, as the BIO strategy ensures that each embedding aggregates information from all other embeddings, resulting in better global perception performance. Additionally, our BIO strategy outperforms the other ordering strategies by a significant margin, indicating its effectiveness.

Finally, we verify the importance of our importance-aware pooling operation (IAP), as shown in Fig.~\ref{fig_ablation2} (b). We compare our IAP with several different pooling strategies, including the average pooling strategy, the max pooling strategy, and the weighted sum strategy. It can be inferred from the figure that preventing the model from being influenced by useless group embeddings introduces a performance improvement of 0.6\%, compared to the direct weighted sum strategy, indicating the effectiveness of our IAP strategy.


\section{Conclusion}
\label{conclusion}
In this work, we introduce PoinTramba, a pioneering hybrid framework that combines the powerful modeling capacity of Transformers with the computational efficiency of Mamba for point cloud analysis. By integrating these two architectures, PoinTramba achieves a superior balance between computational complexity and analytical performance. Additionally, our innovative BIO strategy significantly exploits Mamba's performance.
Extensive experiments demonstrate the effectiveness of PoinTramba, establishing a new state-of-the-art benchmark in point cloud analysis. This novel approach leverages the combined strengths of Transformer and Mamba architectures, marking a significant advancement in the field.
For future work, we plan to explore further optimization techniques to enhance the scalability of PoinTramba and investigate its applicability to a broader range of point cloud tasks and further refine our ordering strategy to further improve efficiency and performance.

\textbf{Limitation.} In this study, we focused solely on an importance-aware ordering strategy. However, it is not clear whether this approach is the most optimal sorting algorithm that can fully harness the potential of Mamba. Additionally, further experiments across a broader range of tasks are necessary to comprehensively evaluate the capabilities of our PoinTramba.


\bibliographystyle{unsrt}
\bibliography{main}

\newpage
\appendix

\section{More Details of Experimental Protocols}
\label{appendix_exp_setting}
\subsection{Datasets}
\label{dataset}
Following previous works~\cite{PointMamba}, we evaluate the effectiveness of our method on three downstream tasks, \ie, real-world object classification on ScanObjectNN~\cite{ScanObjNN}, synthetic object classification on ModelNet40~\cite{ModelNet}, and part segmentation on ShapeNetPart~\cite{ShapeNetPart}. 
The ScanObjNN dataset~\cite{ScanObjNN} is a challenging point cloud object classification dataset consisting of 15,000 point cloud objects from 15 categories scanned from the real world, which includes three variants: OBJ\_BG, OBJ\_ONLY, and PB\_T50\_RS. Following previous works~\cite{PointMamba, PCMamba}, we set the number of points for each sample as 2048.
The ModelNet dataset~\cite{ModelNet} is a widely-used point cloud object classification dataset covering 40 categories where each category includes 100 synthetic samples. Following previous works~\cite{PointMamba, PCMamba}, we set the number of points for each sample as 1024.
The ShapeNetPart~\cite{ShapeNetPart} dataset is a challenging part segmentation dataset for point cloud, consisting of 16,880 samples from 16 different shape categories with 50 part labels. Following previous works~\cite{PointMamba, PCMamba}, we set the number of points for each sample as 2048.

\subsection{Implementation Details}
\label{implement}
For all of our experiments, we adopt AdamW~\cite{AdamW} optimizer with a cosine scheduler. We train our model for 300 epochs, including a warm-up stage of 10 epochs. The initial learning rate is set to 3e-4 for all classification models and 2e-4 for part segmentation models, with a weight decay of 5e-2 for all experiments. The batch size is set to 32 for all tasks.
Our intra-group Transformer encoder includes 4 Transformer layers with a channel size of 384, and no classification token is utilized. Our inter-group Mamba encoder includes 12 Mamba layers with a channel size of 384, and no classification token is utilized. For classification tasks on both the ScanObjectNN and ModelNet datasets, the point cloud samples are divided into 256 groups, with each group consisting of 16 points. For part segmentation tasks on the ShapeNetPart dataset, the point cloud samples are divided into 128 groups, with each group consisting of 32 points. 
The projection layers, including the embedding projection layer and the global projection layer, are two-layer convolutional layers with a hidden dimension of 128 and an output dimension of 256, followed by a BatchNorm layer, a ReLU function, and a normalization operation. The importance score prediction head is a two-layer convolutional layer with a hidden dimension of 128 and an output dimension of 1, followed by a BatchNorm layer and a ReLU function. The hyper-parameters $\alpha$, $\beta$ and $\gamma$ are all set to 1.0 for classification tasks and set to 5.0, 1.0, and 1.0 for part segmentation tasks. Most of our experimental settings are adopted from PointMamba~\cite{PointMamba} including the classification head and segmentation head. All of our classification models are trained on a server with ten NVIDIA RTX 3090 GPUs, and all of our segmentation models are trained on a server with eight NVIDIA A100 GPUs with 80GB memory.

\section{Additional Ablation Study}

\begin{table}[ht]
\caption{Ablation on the sensitivity of our PoinTramba to the number of groups (Group Number) and the number of points within a point group (Group Size). Experiments are conducted on the PB-T50-RS variant of the ScanObjNN dataset.}
\centering
\scalebox{0.9}{
\begin{tabular}{c c c | c c c}
\hline
Group Number & Group Size & Acc. (\%) $\uparrow$ & Group Number & Group Size & Acc. (\%) $\uparrow$ \\
\hline
8 & 1024 & 87.8 & 32 & 256 & 88.3 \\
8 & 512 & 87.8 & 32 & 128 & 88.5 \\
16 & 512 & 89.0 & 64 & 64 & 88.2 \\
16 & 256 & 89.1 & 128 & 32 & 87.3 \\
\hline
\end{tabular}
}
\label{table_group}
\end{table}

In this section, we perform additional ablation studies to verify the effectiveness of our method. First, we investigate the sensitivity of our method to the number of groups $G$ and the number of points within a point group $K$. In particular, we divide the point clouds into several groups ranging from 32 to 1024, with the number of points within each point group varying from 128 to 8. The results are reported in Table~\ref{table_group}. As shown in the table, a group size that is too small results in limited semantic information for each point group. Consequently, the intra-group Transformer encoder may fail to encode the potential semantic features of each point group, leading to limited recognition performance of PoinTramba. Conversely, when the group size is too large, a lightweight intra-group Transformer encoder, \ie, a 4-layer Transformer encoder, may fail to encode abundant features for each point group, also leading to limited recognition performance. Notably, due to the extraordinary long-range modeling capacity of Mamba, varying the number of point groups does not significantly affect the performance of our PoinTramba model.

\begin{table}[ht]
\caption{Ablation study on the sensitivity of our PoinTramba to the number of layers $T$ in the Transformer encoder. Experiments are conducted on the PB-T50-RS variant of the ScanObjNN dataset.}
\centering
\scalebox{0.9}{
\begin{tabular}{l c c c}
\hline
Method & Layer Num. & Acc. (\%) $\uparrow$ & Param. (M) $\downarrow$ \\
\hline
PoinTramba (Ours) & 0 & 87.9 & 12.5 \\
PoinTramba (Ours) & 1 & 87.9 & 14.2 \\
PoinTramba (Ours) & 2 & 88.2 & 16.0 \\
PoinTramba (Ours) & 4 & 89.1 & 19.5 \\
\hline

\end{tabular}
}
\label{table_layer}
\end{table}

Finally, we examine the sensitivity of our PoinTramba to the number of layers $T$ in the Transformer encoder. The results are listed in Table~\ref{table_layer}. As inferred from the table, when the number of layers is small, \eg, 1 or 2, the intra-group Transformer encoder is too lightweight to capture abundant semantic features for each point group. Following this configuration, PoinTramba does not significantly outperform the PoinTramba model that uses MLPs as the intra-group encoder, \ie, when $T = 0$. However, when a four-layer Transformer encoder is used, the intra-group encoder effectively models long-range dependencies within each point group, resulting in outstanding recognition performance. 

\section{Visualization}
\label{vis}

\begin{figure}[h]
\centering
\includegraphics[width=0.95\linewidth]{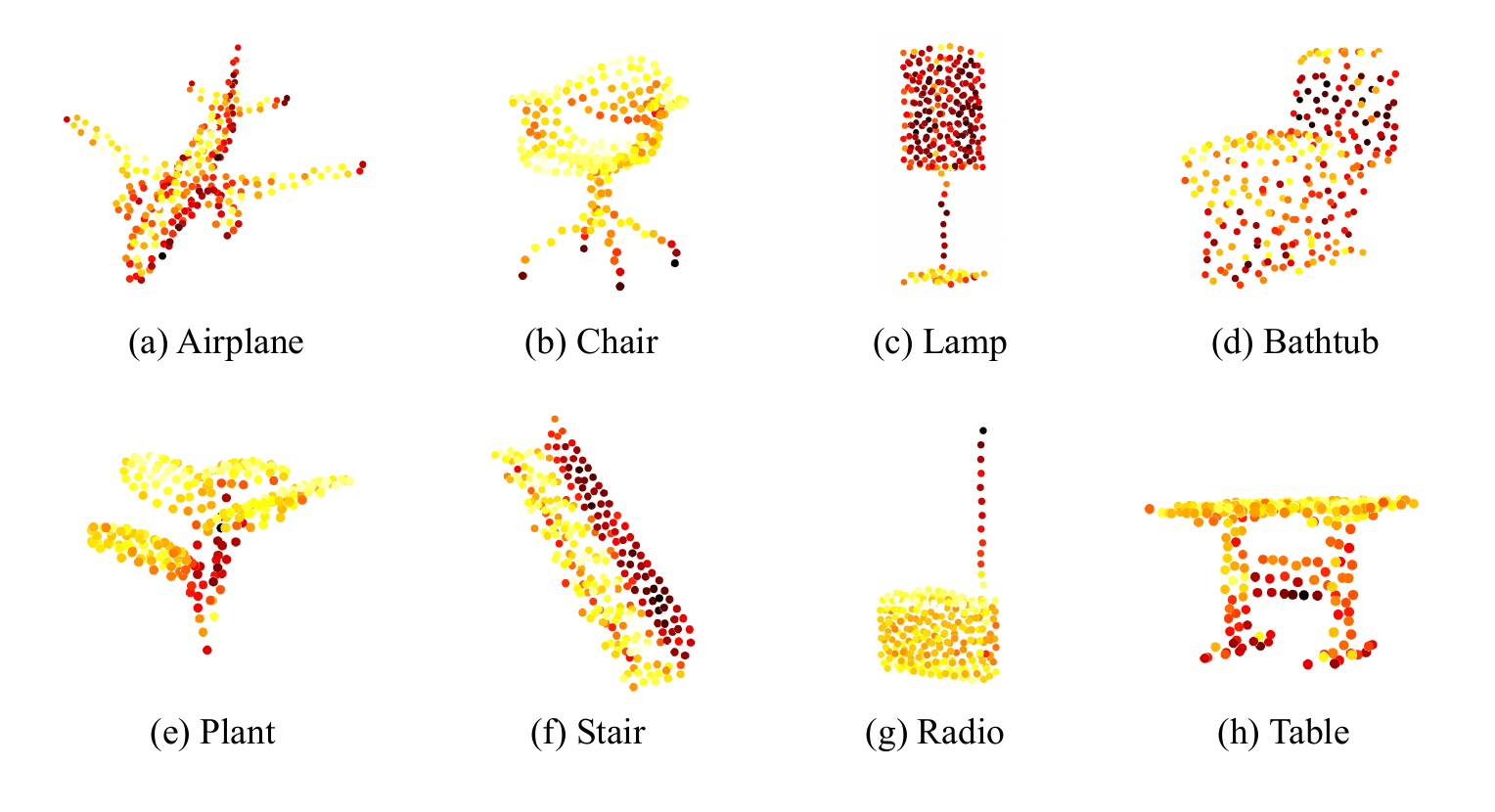}
\caption{Visualization of the importance scores for the point groups predicted by our PoinTramba model. Samples from various categories in ModelNet40 are used as examples. Red regions indicate higher importance scores, while yellow regions indicate lower importance scores.}
\label{fig_vis_importance}
\end{figure}

In this section, we present visualizations of the importance scores predicted by our PoinTramba model in Fig.\ref{fig_vis_importance}. As illustrated in the figure, points associated with the fuselages have higher importance scores than those on the wings. Similarly, points located on the legs of chairs or tables, lampshades, bathtub water tanks, plant stems, stair handrails, and radio antennas also exhibit higher importance scores. Unlike previous conventional ordering strategies\cite{PointMamba, PCMamba}, our BIO strategy reorders the group embeddings in a bi-directional manner based on the learned importance scores. This approach allows the Mamba model to better process structured data, resulting in more refined global inter-group features and significantly enhancing analytical performance.


\end{document}